\documentclass[11pt]{article}

\usepackage[preprint]{acl}

\usepackage{times}
\usepackage{latexsym}

\usepackage[T1]{fontenc}

\usepackage[utf8]{inputenc}

\usepackage{microtype}

\usepackage{inconsolata}

\usepackage{graphicx}
\usepackage{booktabs}
\usepackage{multirow}
\usepackage{makecell}
\usepackage{todonotes}
\usepackage{subfigure}
\usepackage{amssymb}
\usepackage{pgfplots,pgfplotstable}
\tikzset{
  font={\fontsize{8pt}{10}\selectfont}}

\usepackage{hyperref}
\usepackage[table]{xcolor}
\definecolor{lightblue}{RGB}{66,133,244}

\usepackage{dblfloatfix} 
\usepackage{placeins}
\usepackage{multicol}

\newcommand{\simulstream}[0]{\texttt{simulstream}}

\title{Simulstream: Open-Source Toolkit for Evaluation and Demonstration \\ of Streaming Speech-to-Text Translation Systems}

\author{Marco Gaido, Sara Papi, Mauro Cettolo, Matteo Negri, Luisa Bentivogli \\
  Fondazione Bruno Kessler, Trento, Italy \\
  \texttt{\{mgaido, spapi, cettolo, negri, bentivo\}@fbk.eu} \\}

\begin{document}
\maketitle
\begin{abstract}

Streaming Speech-to-Text Translation (StreamST) requires producing translations concurrently with incoming speech under strict latency constraints, demanding models that balance low latency with high translation quality. Despite rapid progress, evaluation remains fragmented across existing frameworks, which make different assumptions about how systems operate---for example, whether they process continuous speech or short pre-segmented audio, and whether they support output revision (\textit{retranslation}) or not (\textit{incremental}).
For instance, SimulEval, the most widely used framework, supports only incremental decoding, assumes short segmented inputs, and lacks a native support for system demonstrations. 
As a result, comparing systems fairly and consistently across studies remains challenging, with no unified solution for benchmarking and interactive demonstration. To address this gap, we introduce \simulstream{}, the first open-source framework for StreamST evaluation and demonstration. It supports both incremental and re-translation decoding on long-form speech, provides fine-grained logging for quality and latency evaluation, and includes an interactive web interface for real-time visualization and comparison.

\begin{itemize}
    \item[{\raisebox{-0.2\height}{\includegraphics[height=1.2em]{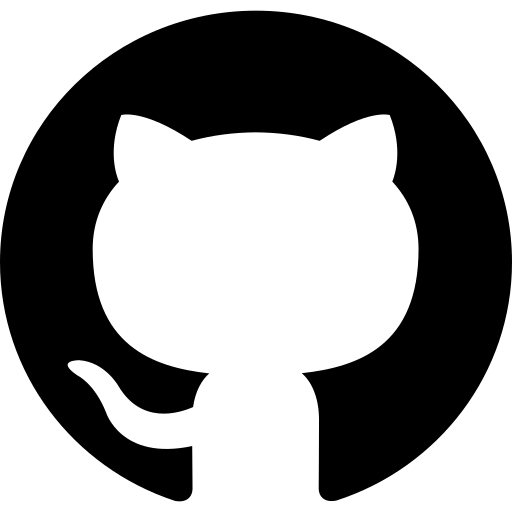}}}] \href{https://github.com/hlt-mt/simulstream}{\texttt{github.com/hlt-mt/simulstream}}
\end{itemize}
\end{abstract}

\section{Introduction}

The task of Streaming Speech-to-Text Translation (StreamST) requires translating continuous, long-form speech input concurrently with the speaker's speech \citep{ren-etal-2020-simulspeech}, posing a significant challenge at the intersection of automatic speech recognition and machine translation \citep{Fugen2008Simultaneous}. Fundamentally distinct from offline translation, where the entire utterance is available before processing, StreamST systems must adhere to specific \textit{latency} constraints, defined as the time elapsed between when a segment of speech is spoken and when its translation is produced. Specifically, translation output must be generated \textit{simultaneously}, or almost in sync, with the incoming speech to remain aligned with the pace of the speaker. This process requires making decisions with only partial information available---often called \textit{policy} \citep{grissom-ii-etal-2014-dont}, which must balance low latency with high-quality translation. Despite these inherent challenges, considerable efforts in both academia and industry have led to rapid progress in the field, as evidenced by the growing number of publications on the topic \citep{papi-etal-2025-real}. 


Two
main paradigms emerged for delivering StreamST outputs: \textit{retranslation} and \textit{incremental} decoding. Retranslation entails continuously re-generating the entire target sentence (or part of it) whenever new source speech is received. This strategy utilizes a non-monotonic decoding approach, enabling dynamic self-correction and revision of previous outputs to maximize translation quality, but often leading to visible output instability,\footnote{\href{https://research.google/blog/stabilizing-live-speech-translation-in-google-translate/}{https://research.google/blog/stabilizing-live-speech-translation-in-google-translate/}} known as flickering \citep{9054585}. Incremental decoding instead does not allow the revision or deletion of previously generated outputs 
(reflecting an ``append-only'' approach)
at the benefit of a more stable user experience \citep{dalvi-etal-2018-incremental}. Retranslation is prevalent in industrial applications \citep{arivazhagan-etal-2020-translation} as it is easy to apply, requiring no specific modification or re-training to the underlying offline speech translation model. In contrast, research has largely focused on incremental methods, including both training-based \citep{schneider-waibel-2020-towards,IRANZOSANCHEZ2021303,ouyang-etal-2025-infinisst}, and training-free approaches \citep{10832264,papi-etal-2024-streamatt}. 
As both paradigms mature, understanding their trade-offs under consistent conditions has become essential for progress in simultaneous speech translation.

However, systematic comparison between these paradigms remains an open challenge. Existing frameworks typically support only subsets of the StreamST pipeline, preventing fair and reproducible benchmarking across decoding strategies. The most widely used toolkit, SimulEval \citep{ma-etal-2020-simuleval}, was only designed for incremental decoding and does not support token deletion, making it incompatible with retranslation methods. In addition, it assumes short speech segments rather than continuous long-form streams, which limits its applicability to modern StreamST benchmarks \citep{agostinelli-etal-2025-findings,polak2025better}. It also lacks support for interactive system demonstrations and is no longer actively maintained.\footnote{SimulEval has been archived on September 18th, 2025.} Beyond SimulEval, only a limited number of alternative tools have been proposed, with scarce adoption and providing only partial solutions: SLTev \citep{ansari-etal-2021-sltev} supports both paradigms but lacks demonstration capabilities, while Lecture Translator \citep{huber-etal-2023-end} offers a web interface without an open evaluation backend. As a result, the community lacks a unified, maintained framework for both evaluating and demonstrating StreamST systems.

To address 
this gap,
we propose \simulstream, the first open-source (Apache 2.0 License) tool for unified evaluation and demonstration of StreamST systems. 
Our contributions are as follows:


\begin{itemize}
\item We introduce \simulstream{}, the first unified open-source framework for long-form generation and evaluation of StreamST systems under both incremental and re-translation strategies, with accurate tracking of emitted and deleted tokens.
\item We provide a lightweight proxy interface to the community-standard SimulEval framework, supporting direct reuse of existing systems without modification.
\item We release an interactive web-based demonstration platform for real-time visualization and system comparison.\footnote{Demo: \url{https://youtu.be/k0WVOIMZFY8}}
\end{itemize}

Beyond enabling reproducible benchmarking, \simulstream{} establishes a common infrastructure for studying and comparing different StreamST decoding paradigms under controlled and consistent conditions. This unified setup has supported its adoption as the official evaluation toolkit for the Simultaneous Translation track of the IWSLT 2026 Evaluation Campaign \citep{adelani-etal-2026-speech}.



\section{The \simulstream{} Tool}




\simulstream{} provides a WebSocket server and supporting utilities for running streaming speech processing experiments and demonstrations. It supports real-time transcription and translation over streaming audio input. In this context, streaming assumes a continuous, unbounded speech signal, rather than a sequence of short, pre-segmented utterances as in earlier simultaneous speech processing evaluations (e.g., SimulEval). The short-form setting can be emulated by splitting the input audio into small consecutive segments.

\begin{figure}[!ht]
    \centering
    \includegraphics[width=0.95\linewidth]{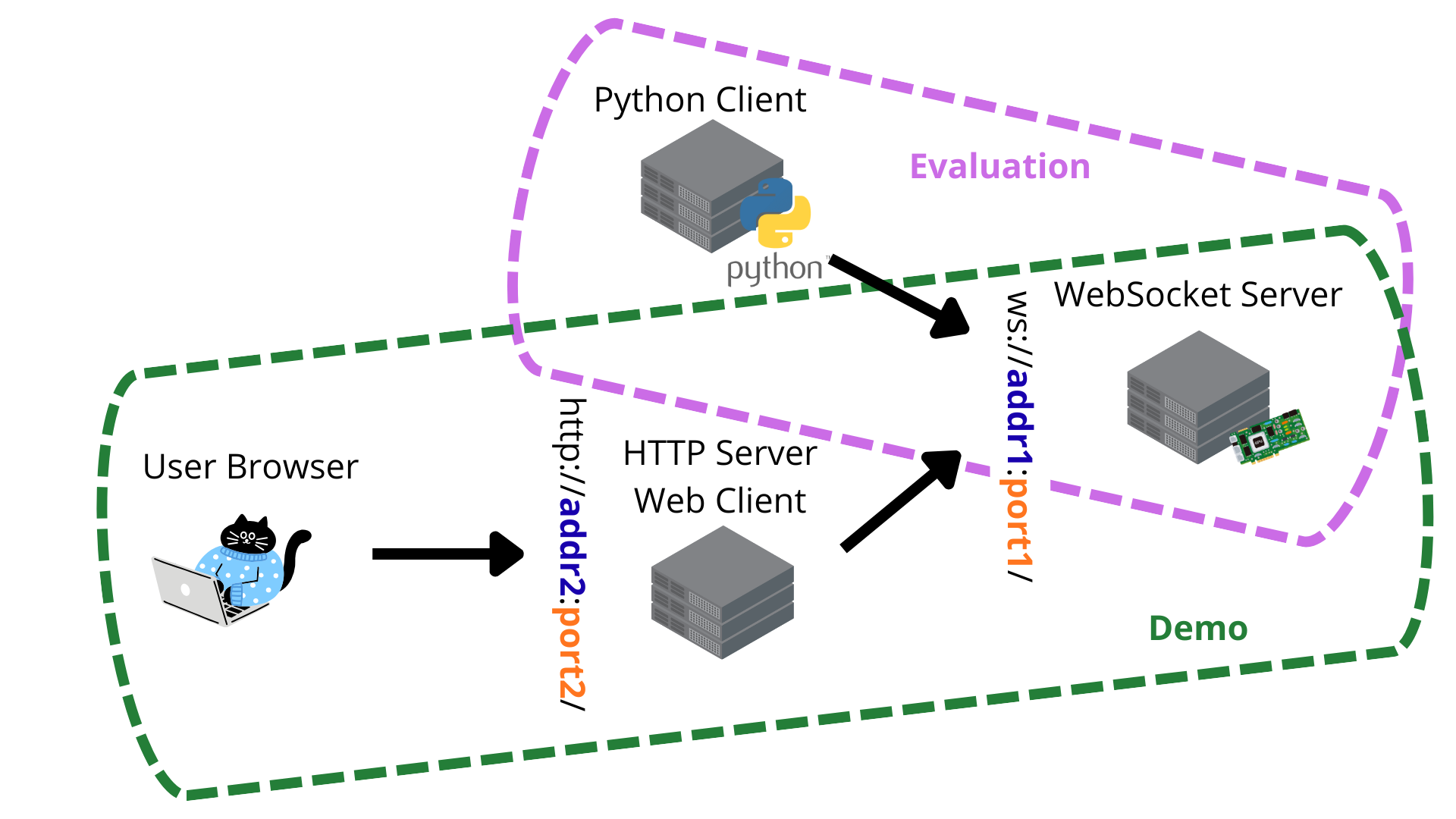}
    \caption{Architecture of the \simulstream{} tool.}
    \label{fig:architecture}
\end{figure}

\subsection{Architecture}

The package is based on a full-duplex WebSocket client-server interaction, shown in Figure \ref{fig:architecture}. The server waits for connections from the clients. The clients can send configuration messages in JSON format (e.g., to specify the input/output desired 
languages)
and audio chunks, encoded as 16-bit integers. The server processes the input audio and sends to the client JSONs that contain the transcript/translation generated by the configured \textit{speech processor}, which implements the logic to generate outputs from incoming audio, as in SimulEval agents. In addition, if configured to do so, the server writes metric logs into a JSONL file, reporting generated (and deleted) text at each step together with computational costs and total audio processed, which can be used to compute metrics.

The codebase is organized into 3 main blocks: the server, the clients, and the evaluation 
(see \S\ref{sec:eval}). 

\paragraph{Server.}
The WebSocket server is launched via the \texttt{simulstream\_server} command, which accepts two YAML configuration files. The first defines server-level settings (e.g., IP/DNS address and port), while the second specifies the \textit{speech processor} (see \S\ref{sec:speech_proc}). The server creates a fixed-size pool of processors to support concurrent client execution. Incoming connections exceeding the pool capacity are rejected, providing explicit control over concurrency and preventing resource exhaustion and out-of-memory errors under high load.

\begin{figure}
    \centering
    \includegraphics[width=0.97\linewidth]{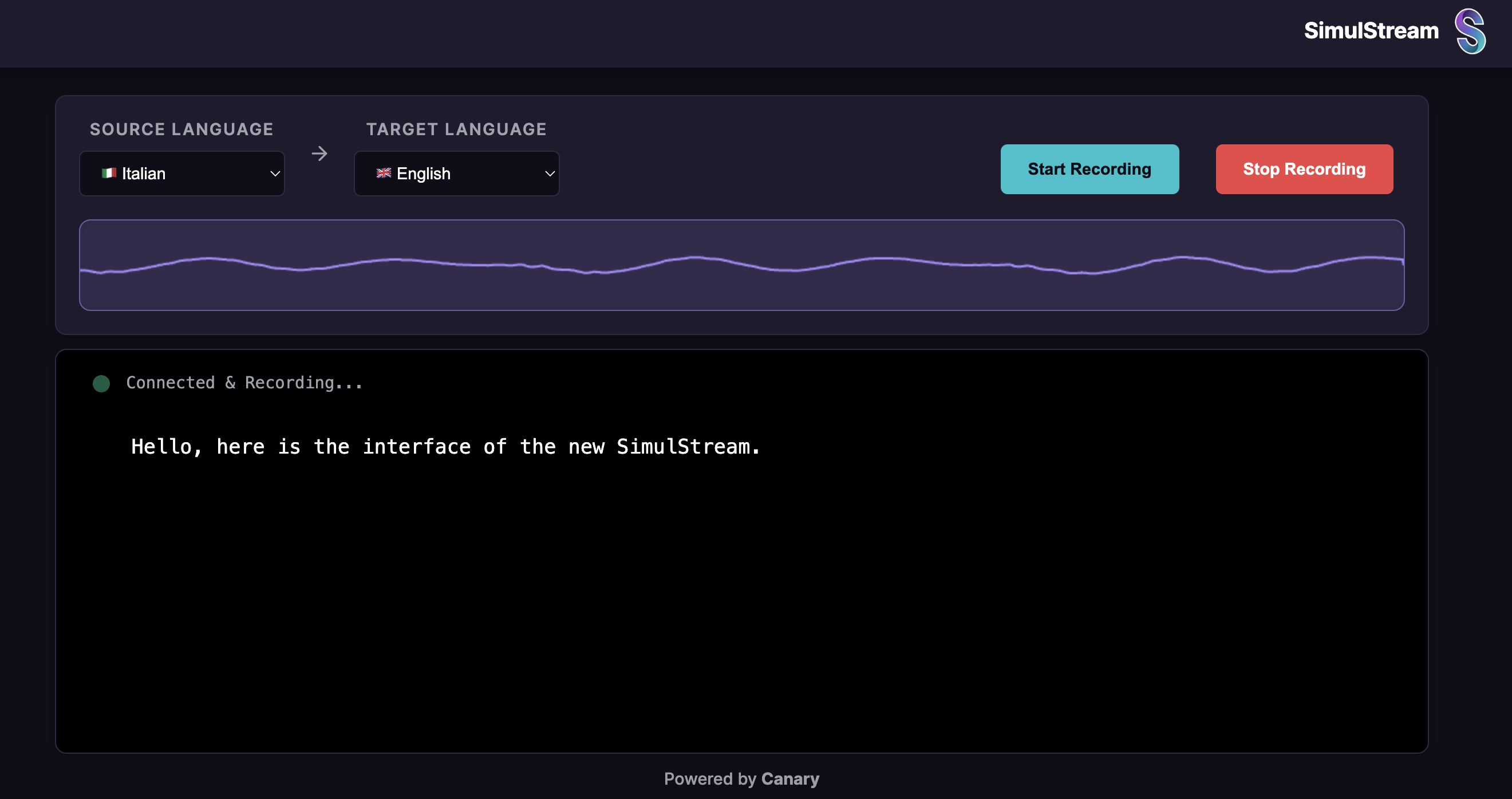}
    \caption{Screenshot of the web interface.}
    \label{fig:web_interface}
\end{figure}

\paragraph{Clients.}
The tool comes with two clients, an HTTP web server and a command-line WAV client, 
meant, respectively, 
for demo and evaluation of a speech processor. The HTTP web server provides an HTML/CSS/Javascript web interface interacting with the WebSocket server (see Figure \ref{fig:web_interface}). The command-line WAV client, instead, gets as argument a file containing a list of WAV files to stream to the server and it can be used to produce the metric logs 
necessary
for evaluating a speech processor. To further ease experimenting and evaluating speech processors, \simulstream{} includes a dedicated command that processes a list of audios without setting up a client-server interaction and writes the metric logs into a specified file.

\subsection{Speech Processors}
\label{sec:speech_proc}

\simulstream{} also comes with several streaming speech processors, which can also be used as examples to build custom ones.

\begin{itemize}
    \item Sliding window \citep{sen2022simultaneoustranslationunsegmentedinput}: a speech processor that applies a fixed-length sliding-window retranslation with deduplication to mitigate overlapping outputs when processing unsegmented audio streams. The approach relies on detecting the longest common subsequence between the current window and the previous one, in order to prevent repeating tokens caused by overlapping audio windows. This approach is implemented for Canary v2 \citep{sekoyan2025canary1bv2parakeettdt06bv3efficient}, Seamless \citep{communication2023seamlessm4tmassivelymultilingual}, and any speech-to-text HuggingFace model.
   
\item StreamAtt \citep{papi-etal-2024-streamatt}: a speech processor that leverages audio-textual alignments based on cross-attention scores. This mechanism serves two purposes: \textit{i)} identifying which part of the generated hypothesis should be emitted (i.e., applying the AlignAtt policy; \citealt{papi23_interspeech}), and \textit{ii)} determining which part of the audio history to retain in memory, based on the textual history (i.e., the previously generated outputs), thereby enabling long-form processing. This approach is implemented for Seamless and Canary, but can be ported to any speech-to-text attention-based HuggingFace model.

    \item SimulEval Agent Wrapper: a wrapper processor that calls a 
    SimulEval-based
    agent.\footnote{It supports agents implemented for SimulEval>=1.1.0.} This enables seamless porting of any existing system developed using the legacy 
    tool.
    \item VAD Wrapper: speech processor that integrates Voice Activity Detection 
    (VAD), specifically the Silero VAD \citep{Silero}. Its function is to split continuous audio streams into speech chunks by filtering out parts that do not contain speech before passing the remaining segments to an underlying speech processor.

\end{itemize}

The repository is structured to be easily extended with custom speech processors. To this aim, users should create a subclass of \texttt{simulstream.server.speech\_processors.\\SpeechProcessor}, add the class to the \texttt{PYTHONPATH}, and reference it in the YAML file configuration of the speech processor. Subclasses must implement methods to load models, process audio chunks, set source/target languages, and clear internal states. Speech chunks are fed as 1D NumPy 
arrays
containing PCM audio normalized to the range $[-1.0, 1.0]$ sampled at 16 kHz. Their size (in seconds) is determined by the speech processor itself through a dedicated method and the speech processor has to return incremental outputs, containing the tokens to be deleted from previously generated outputs (if any) and the new tokens to be emitted.




\subsection{Evaluation}
\label{sec:eval}

The evaluation of StreamST systems is typically 
framed as a trade-off between latency and translation quality.
\simulstream{} provides built-in metrics for both dimensions, while also exposing a simple interface for integrating custom or additional evaluation measures. 
Moreover, evaluations operate directly on fine-grained metric logs that record the full behavior of a speech processor during inference. As a result, new metrics can be computed a posteriori without requiring additional information, and can be applied to any previously executed run.

\paragraph{Quality.}
Most simultaneous and streaming ST
work has relied solely on BLEU \citep{papineni-etal-2002-bleu}, 
despite growing evidence that it is suboptimal compared to modern neural metrics with higher correlation to human judgments
\citep{freitag-etal-2022-results}. 
Accordingly, \simulstream{} implements both BLEU (computed via sacreBLEU \citep{post-2018-call}) and COMET\footnote{The default model is \texttt{Unbabel/wmt22-comet-da}, but alternative models can be specified via a command-line parameter.} \citep{rei-etal-2020-comet}.
The scores are computed after a re-segmentation of the full generated text for an audio to match the segmentation of the references, which are assumed to be segmented at the sentence level. 
This is achieved using
\texttt{mweralign} \citep{post-hoang-2025-effects}, 
a Python 
implementation
of the widespread \texttt{mwerSegmenter} binary executable \citep{matusov-etal-2005-evaluating} adopted in ST evaluation campaigns \citep{agostinelli-etal-2025-findings} and previous work on StreamST \citep{papi-etal-2024-streamatt}. 
For streaming processors that may revise or delete previously emitted tokens (e.g., sliding-window re-translation), evaluation is performed on the final output sequence only, while intermediate revised hypotheses are not considered.

\paragraph{Latency.} \simulstream{} includes a reimplementation of the StreamLAAL metric \citep{papi-etal-2024-streamatt} for latency evaluation. The main difference is the use of \texttt{mweralign} instead of \texttt{mwerSegmenter} for automatic resegmentation of generated text against sentence-level references.
StreamLAAL assumes a monotonic word-level alignment between system output and reference. Latency is then computed as the difference between the emission time of each generated word and its expected reference time, where reference words are assumed to be uniformly distributed over the utterance duration (e.g., for a 5-second utterance with 10 words, each word is assigned a 0.5-second timestamp).
The emission time of a generated word can be defined in two ways: the \textit{ideal} (or \textit{computational-unaware}) time, corresponding to the amount of audio processed when the word is produced, or the \textit{computational-aware} (CA) time, which additionally accounts for model execution time. StreamLAAL refers to the former, while StreamLAAL\textsubscript{CA} denotes the latter.
Since StreamLAAL was originally designed for systems that do not revise previously emitted tokens, we extend its definition to retranslation processors by computing latency on the final output sequence, consistent with the quality evaluation. Specifically, for each word, we use its last update time as the emission time. This design choice may overestimate latency compared to user experience in retranslation systems, while boosting the translation quality as it does not account for intermediate---possibly wrong---outputs. Lastly, for character-level languages such as Chinese, Japanese, and Korean, the latency unit is represented by a single character, similarly to SimulEval.

\paragraph{Other statistics.}
\simulstream{} additionally computes statistics that complement the evaluation metrics defined above. 
Specifically, it 
enables computing \textit{normalized erasure} (NE) and the \textit{real time factor} (RTF). NE \citep{arivazhagan-etal-2020-translation} measures flickering in retranslation. It is defined as the ratio between the number of tokens that have been deleted and the number of final generated tokens. RTF, instead, is measured as seconds spent in computation for each input audio second. 
RTF values
greater than 1 
indicate
that the system is not able to process the input in time before the next input arrives. As for the metrics above, the system modularity enables easy integration of additional/customized statistics.

\begin{table*}[t]
\centering
\footnotesize
\setlength{\tabcolsep}{5.5pt}
\begin{tabular}{lllcclcccc}
\toprule
\multicolumn{2}{c}{\textbf{Speech P.}}  & \textbf{Model} & \textbf{w/f/t} &
\textbf{COMET $\uparrow$} & \textbf{BLEU $\uparrow$} & \textbf{StreamLAAL $\downarrow$} & \textbf{StreamLAAL\textsubscript{CA} $\downarrow$} &
\textbf{NE $\downarrow$} & \textbf{RTF $\downarrow$} \\
\midrule
\multirow{16}{*}{\rotatebox{90}{\textcolor{blue}{\texttt{retranslation}}}} & \multirow{8}{*}{\rotatebox{90}{Sliding Window}} & \multirow{4}{*}{Canary}
 & 8  & 0.7853 & 28.71 & 2.47 & 2.80 & 0.5771 & 0.1438 \\
& & & 10 & 0.7883 & 28.48 & 3.01 & 3.43 & 0.8049 & 0.1865 \\
& & & 12 & 0.7957 & 28.92 & 3.40 & 3.93 & 0.9475 & 0.2315 \\
& & & 14 & \textbf{0.7986} & \textbf{29.20} & 3.93 & 4.47 & 1.1277 & 0.2405 \\
\cmidrule(lr){3-10}
& & \multirow{4}{*}{SeamlessM4T}
 & 8  & 0.7174 & 24.51 & 2.42 & 2.80 & 0.8586 & 0.1770 \\
& & & 10 & 0.7359 & 25.39 & 2.92 & 3.38 & 1.0215 & 0.2101 \\
& & & 12 & 0.7439 & 25.70 & 3.53 & 4.07 & 1.1778 & 0.2432 \\
& & & 14 & 0.7469 & 25.65 & 4.26 & 4.90 & 1.3401 & 0.2878 \\
\cmidrule(lr){2-10}
 & \multirow{8}{*}{\rotatebox{90}{VAD + Sliding Window}} & \multirow{4}{*}{Canary} & 0.6 & 0.7375 & 24.70 & 2.54 & 2.78 & 0.1304 & \textbf{0.0822} \\
& & & 0.5 & 0.7426 & 25.01 & 2.57 & 2.82 & 0.1393 & 0.0834 \\
& & & 0.4 & 0.7464 & 25.26 & 2.57 & 2.83 & 0.1488 & 0.0854 \\
& & & 0.3 & 0.7523 & 25.57 & 2.61 & 2.87 & 0.1631 & 0.0866 \\
 \cmidrule(lr){3-10}
&  & \multirow{4}{*}{SeamlessM4T} & 0.6  & 0.6949 & 21.97 & 2.53 & 2.81 & 0.1631 & 0.0939 \\
& & & 0.5 & 0.6972 & 22.16 & 2.56 & 2.84 & 0.1709 & 0.0949 \\
& & & 0.4 & 0.7008 & 22.49 & 2.63 & 2.92 & 0.1842 & 0.0973 \\
& & & 0.3 & 0.7058 & 22.98 & 2.74 & 3.04 & 0.2060 & 0.1008 \\
\midrule
\multirow{8}{*}{\rotatebox{90}{\textcolor{magenta}{\texttt{incremental}}}} & \multirow{8}{*}{\rotatebox{90}{StreamAtt}} & \multirow{4}{*}{Canary} & 2 & 0.7365 & 17.62 & 2.89 & 3.52 & \textbf{0.0000} & 0.5479 \\ 
& & & 4 & 0.7370 & 17.78 & 2.92 & 3.59 & \textbf{0.0000}  & 0.5791 \\
& & & 6 & 0.7422 & 18.23 & 3.26 & 4.04 & \textbf{0.0000} & 0.6315 \\
& & & 8 & 0.7452 & 18.26 & 3.53 & 4.21 & \textbf{0.0000} & 0.5162 \\
 \cmidrule(lr){3-10}
 & & \multirow{4}{*}{SeamlessM4T}
 & 2 & 0.7589 & 26.42 & \textbf{1.95} & \textbf{2.23} & \textbf{0.0000} & 0.2196 \\ 
& & & 4 & 0.7653 & 27.25 & 2.25 & 2.53 & \textbf{0.0000}  & 0.2328 \\
& & & 6 & 0.7695 & 27.80 & 2.56 & 2.85 & \textbf{0.0000} & 0.2447 \\
& & & 8 & 0.7696 & 27.94 & 2.84 & 3.13 & \textbf{0.0000} & 0.2581 \\ 
\bottomrule
\end{tabular}
\caption{Quality (COMET, BLUE), latency (StreamLAAL, StreamLAAL\_CA), flickering (NE), and computational cost (RTF) of the supported \textcolor{blue}{retranslation} and \textcolor{magenta}{incremental} speech processors (\textbf{Speech P.}) with Canary and SeamlessM4T v1 medium on the MuST-C test set (averaged over all language pairs). The column w/f/t refers to varying, the window length (w) for sliding window, the number of frames (f) for StreamAtt, and the VAD probability threshold (t) for VAD-based sliding window.}
\label{tab:avg_all}
\end{table*}

\section{Experiments}

\subsection{Setup}

We demonstrate the utility of our tool by comparing the released speech processors (\S\ref{sec:speech_proc}). Specifically, for both Canary v2 and Seamless medium v1,\footnote{We also experimented with Seamless large v2 but obtained inferior results (see Appendix \ref{app:seamless2large}).} we include \textit{i)} retranslation sliding window, \textit{ii)} the VAD wrapper with retranslation sliding window as underlying processor, and \textit{iii)} incremental StreamAtt. For the sliding window and VAD wrapper, we slide the window by 2s.\footnote{We tested 1s, 2s, and 3s on the dev set and chose 2s as it gave the best quality/latency trade-off.} For the sliding window approach, we vary the window length (8, 10, 12, 14 seconds) to obtain different operating points in terms of quality and latency and maintain the VAD probability threshold fixed at 0.1. For the VAD wrapper, instead, we use 14s as window lenght and we vary the VAD threshold from 0.3 to 0.6, which controls whether to be more conservative (lower values) or more aggressive (higher values) in considering audio as non-speech and filtering it out. Lastly, for StreamAtt, we use speech chunks of 1s, and we vary the cutoff frame with 2, 4, 6, and 8 to get different latency-quality trade-offs. A single NVIDIA A40 40GB is used for all the inferences.

As data, we leverage the 8 language pairs (en$\rightarrow$de,es,fr,it,nl,pt,ro,ru) of the widespread MuST-C dataset \citep{di-gangi-etal-2019-must} and we process the entire TED talks in the test set 
(each lasting $\sim$10 minutes). As an additional test set, we employ MCIF \citep{papi2026mcif}, consisting of  scientific presentations of $\sim$5-6 minutes each covering en$\rightarrow$de,it,zh
and featuring additional difficulties compared to MuST-C including different recording qualities, accents, and domain-specific terminology.
We do not rely on popular ST corpora, e.g., CoVoST2 \citep{wang2020covost} and FLEURS \citep{conneau2023fleurs}, 
as 
they contain sentence-level audios (lasting few seconds), which are not representative of real use cases of long audio streams.



    



\pgfplotstableread[row sep=\\]{
COMET	BLUE  LAAL  LAALCA \\
0.6014	18.79	3.07	3.43 \\
0.6146	18.07	3.80	4.25 \\
0.6183	17.76	4.47	5.01 \\
0.6161	16.16	5.58	6.23 \\
}\seamlessretranslationde

\pgfplotstableread[row sep=\\]{
COMET	BLUE  LAAL  LAALCA \\
0.6535	19.48	1.49	1.67 \\
0.6723	22.17	1.78	1.98 \\
0.6799	22.86	1.96	2.16 \\
0.6859	23.82	2.31	2.51 \\
}\seamlessincrementalde

\pgfplotstableread[row sep=\\]{
COMET	BLUE  LAAL  LAALCA \\
0.6746	17.43	2.97	3.52 \\
0.6814	17.40	3.07	3.72 \\
0.6898	17.99	3.53	4.12 \\
0.6906	18.01	3.84	4.54 \\
}\canaryincrementalde

\pgfplotstableread[row sep=\\]{
COMET	BLUE  LAAL  LAALCA \\
0.6855	26.01	2.68	2.99 \\
0.7071	25.67	3.21	3.66 \\
0.7490	30.19	3.52	4.08 \\
0.7593	30.02	4.06	4.70 \\
}\canaryretranslationde

\pgfplotstableread[row sep=\\]{
COMET	BLUE  LAAL  LAALCA \\
0.7046	31.38	3.07	3.40 \\
0.7137	31.06	3.65	4.07 \\
0.7195	30.61	4.55	5.04 \\
0.7140	29.06	5.49	6.05 \\
}\seamlessretranslationit

\pgfplotstableread[row sep=\\]{
COMET	BLUE  LAAL  LAALCA \\
0.7198	27.59	1.47	1.64 \\
0.7420	31.84	1.68	1.86 \\
0.7496	33.56	1.90	2.09 \\
0.7585	34.79	2.22	2.41 \\
}\seamlessincrementalit

\pgfplotstableread[row sep=\\]{
COMET	BLUE  LAAL  LAALCA \\
0.7389	23.55	2.88	3.54 \\
0.7448	25.71	2.84	3.47 \\
0.7402	24.70	2.10	2.86 \\
0.7614	26.60	3.11	3.64 \\
}\canaryincrementalit

\pgfplotstableread[row sep=\\]{
COMET	BLUE  LAAL  LAALCA \\
0.7811	37.96	2.56	2.90 \\
0.7888	37.59	3.03	3.40 \\
0.8145	43.18	3.43	3.94 \\
0.8186	43.81	3.80	4.38 \\
}\canaryretranslationit

\pgfplotstableread[row sep=\\]{
COMET   BLEU	LAAL  LAALCA \\
0.6295	25.02	3.29	3.59 \\
0.6234	24.62	4.14	4.51 \\
0.6335	24.87	5.17	5.59 \\
0.6307	23.13	6.33	6.85 \\
}\seamlessretranslationzh

\pgfplotstableread[row sep=\\]{
COMET   BLEU	LAAL  LAALCA \\
0.6845	27.35	3.18	3.50 \\
0.6957	28.78	3.41	3.68 \\
0.6992	28.78	3.64	3.91 \\
0.6978	28.85	3.39	3.64 \\
}\seamlessincrementalzh

\begin{figure*}[!ht]
\centering
\small
\subfigure[en-de]{
\begin{tikzpicture}
    \begin{axis}[
            ymajorgrids=true,
            xtick pos=left,
            ytick pos=left,
            minor y tick num=1,
            minor x tick num=1,
            ymin=0.59,
            ymax=0.775,
            xmin=1.25,
            xmax=6.5,
            ylabel=COMET, xlabel=StreamLAAL (s),
            ylabel shift={-4pt},
            width=5.8cm,
            height=6.2cm,
            xtick=data,
            compat=newest,
            xtick={1,2,3,4,5,6},
            ytick={0.6,0.65,0.7,0.75},
            every axis plot/.append style={thick},
            legend style={at={(0.5,-0.2)},    
                    anchor=north,legend columns=3},
        ]
        \addplot[color=blue, mark=*] table[x=LAAL,y=COMET]{\seamlessretranslationde};
        \addplot[dashed, color=blue, mark=*] table[x=LAALCA,y=COMET]{\seamlessretranslationde};
        \addplot[color=magenta, mark=*] table[x=LAAL,y=COMET]{\seamlessincrementalde};
        \addplot[dashed, color=magenta, mark=*] table[x=LAALCA,y=COMET]{\seamlessincrementalde};

        \addplot[color=magenta, mark=square*] table[x=LAAL,y=COMET]{\canaryincrementalde};
        \addplot[color=magenta, mark=square*, dashed] table[x=LAALCA,y=COMET]{\canaryincrementalde};
        \addplot[color=blue, mark=square*] table[x=LAAL,y=COMET]{\canaryretranslationde};
        \addplot[dashed, color=blue, mark=square*] table[x=LAALCA,y=COMET]{\canaryretranslationde};
    \end{axis}
\end{tikzpicture}
}
\subfigure[en-it]{
\begin{tikzpicture}
    \begin{axis}[
            ymajorgrids=true,
            xtick pos=left,
            ytick pos=left,
            minor y tick num=1,
            minor x tick num=1,
            ymin=0.70,
            ymax=0.825,
            xmin=1.25,
            xmax=6.5,
            xlabel=StreamLAAL (s),
            ylabel shift={-4pt},
            width=5.8cm,
            height=6.2cm,
            xtick=data,
            compat=newest,
            xtick={1,2,3,4,5,6},
            every axis plot/.append style={thick},
            legend style={at={(0.425,1.2)},    
                    anchor=north,legend columns=3},
        ]
        \addplot[color=blue, mark=*] table[x=LAAL,y=COMET]{\seamlessretranslationit};
        \addplot[dashed, color=blue, mark=*] table[x=LAALCA,y=COMET]{\seamlessretranslationit};
        \addplot[color=magenta, mark=*] table[x=LAAL,y=COMET]{\seamlessincrementalit};
        \addplot[dashed, color=magenta, mark=*] table[x=LAALCA,y=COMET]{\seamlessincrementalit};

        \addplot[color=magenta, mark=square*] table[x=LAAL,y=COMET]{\canaryincrementalit};
        \addplot[color=magenta, mark=square*, dashed] table[x=LAALCA,y=COMET]{\canaryincrementalit};
        \addplot[color=blue, mark=square*] table[x=LAAL,y=COMET]{\canaryretranslationit};
        \addplot[dashed, color=blue, mark=square*] table[x=LAALCA,y=COMET]{\canaryretranslationit};        
        \legend{Re-translation, , Incremental}
    \end{axis}
\end{tikzpicture}
}
\subfigure[en-zh]{
\begin{tikzpicture}
    \begin{axis}[
            ymajorgrids=true,
            xtick pos=left,
            ytick pos=left,
            minor y tick num=1,
            minor x tick num=1,
            ymin=0.62,
            ymax=0.71,
            xmin=3,
            xmax=7,
            xlabel=StreamLAAL (s),
            ylabel shift={-4pt},
            width=5.8cm,
            height=6.2cm,
            xtick=data,
            compat=newest,
            xtick={1,2,3,4,5,6,7},
            every axis plot/.append style={thick},
            legend style={at={(0.0,1.2)},    
                    anchor=north,legend columns=3},
        ]
        \addplot[color=blue, mark=*] table[x=LAAL,y=COMET]{\seamlessretranslationzh};
        \addplot[dashed, color=blue, mark=*] table[x=LAALCA,y=COMET]{\seamlessretranslationzh};
        \addplot[color=magenta, mark=*] table[x=LAAL,y=COMET]{\seamlessincrementalzh};
        \addplot[dashed, color=magenta, mark=*] table[x=LAALCA,y=COMET]{\seamlessincrementalzh};
    \end{axis}
\end{tikzpicture}
}
\caption{Latency (StreamLAAL$\downarrow$) - Quality (COMET$\uparrow$) curves of Sliding-window \textcolor{blue}{retranslation} and StreamAtt \textcolor{magenta}{incremental} methods applied to SeamlessM4T v1 medium ({\Large$\bullet$}) and Canary v2 ($\blacksquare$) on the three target languages of MCIF (except for Canary, which does not support Chinese). Dashed lines indicate computationally aware latency, while solid lines computationally unaware. Numerical results are presented in Appendix \ref{app:mcif-num-results}.}
\label{fig:simul_res}
\end{figure*}
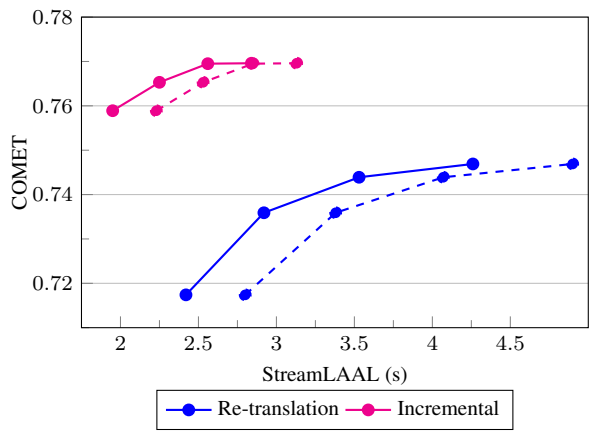

\subsection{Results}

Table \ref{tab:avg_all} reports the scores obtained by the different speech processors on MuST-C averaged over all the 8 language pairs.\footnote{Results for each language pair are available at \url{https://github.com/hlt-mt/simulstream/blob/main/examples/must_results/README.md}.} By comparing the retranslation approaches with the two models (Canary and SeamlessM4T), Canary emerges as the best performing one by a large margin with analogous trends both with and without VAD. Although their latency (StreamLAAL) is similar, the quality (COMET, SacreBLEU) of Canary's translations is significantly better 
than that of
SeamlessM4T. Canary also displays a lower amount of flickering (NE) and computational costs (RTF).

Moving to the comparison between the pure sliding window approach and the one preceded by the VAD wrapper, the behavior is consistent for both models. The flickering and computational costs are dramatically reduced (flickering is roughly 8$\times$ smaller, and RTF 2-3$\times$ lower). On the downside, there is 
a reduced possibility of controlling
the latency-quality tradeoff (with the VAD wrapper, all the operational points are close to each other, with latency differences spanning 0.2s and quality less than 0.015 COMET) and, most importantly, the translation quality suffers a huge drop.


Lastly, we compare incremental decoding with StreamAtt against retranslation using the best performing sliding-window strategy. To the best of our knowledge, this is the first direct comparison between incremental and retranslation approaches, made possible by the unified implementation and evaluation framework provided by \simulstream{}. Notably, although our evaluation metrics are designed to penalize latency while favoring the quality of retranslation methods (see \S\ref{sec:eval}), StreamAtt on Seamless surprisingly outperforms the sliding-window approach not only in latency but also in translation quality, while completely eliminating flickering effects. The main drawback is its higher computational cost (RTF), particularly in low-latency settings, although still far from reaching the critical value of 1.0. In contrast, Canary exhibits the opposite behavior, with the sliding window achieving substantial quality gains at comparable or even lower latency than StreamAtt.

The results on MCIF (Figure~\ref{fig:simul_res}) further confirm this trend: retranslation consistently achieves the highest translation quality with Canary, whereas StreamAtt yields the lowest latency with Seamless. 
%
%
Overall, no single decoding strategy emerges as universally superior across models, indicating that the optimal quality--latency trade-off is inherently model- and application-dependent.\footnote{For instance, incremental decoding remains essential for applications such as cascaded speech-to-speech translation systems \citep{1620853.1620895,sudoh2020simuls2s}.} These findings highlight the importance of \simulstream{} as a unified framework for fair, systematic, and reproducible comparison of simultaneous decoding strategies, supporting informed design choices for future streaming speech translation systems.






\section{Conclusions}

We presented \simulstream, the first open-source framework enabling unified evaluation and demonstration of Streaming Speech-to-Text Translation (StreamST) systems. Unlike previous tools, \simulstream{} supports both incremental decoding and retranslation, tracks token emissions and deletions, and accommodates long-form audio, providing a comprehensive platform for assessing translation quality, latency, and computational costs. In addition, its interactive web interface allows real-time visualization and comparison of different StreamST systems, bridging the gap between research and practical deployment. By offering a paradigm-agnostic, flexible, and extendable toolkit, \simulstream{} empowers the community to systematically study, benchmark, and showcase modern StreamST approaches, facilitating the advancement of simultaneous ST research.

\newpage
\section*{Acknowledgments}

We acknowledge the support of the project InnovAction: Network Italiano dei Centri per l’Innovazione Tecnologica (CUP B47H2200437000), funded by MIMIT with NPRR
- NextGenerationEU funds, in collaboration with Piazza Copernico S.r.l. 
This paper has received funding from the European Union’s Horizon Europe programme grant agreement No. 101213369 (project DVPS).


\bibliography{custom}
\clearpage
\appendix

\onecolumn
\begin{multicols}{2}
\section{Comparison with SeamlessM4T v2 Large}
\label{app:seamless2large}

Table \ref{tab:seamless_comp} reports the scores for SeamlessM4T v2 large and SeamlessM4T v1 medium. Interestingly, SeamlessM4T v1 medium outperforms by a large margin the v2 large version on all metrics -- including those related to output quality -- but normalized erasure. The latency is higher not only due to the (expected) higher computational cost, evident from the high RTF, but also the ideal StreamLAAL is notably worse. For this reason, the results in the main paper are reported using the medium v1 version.

\columnbreak

\section{Numerical Results on MCIF}
\label{app:mcif-num-results}

Numerical results for Figure \ref{fig:simul_res} are presented in Table \ref{tab:mcif-results} in the next page.

\end{multicols}

\begin{table*}[!t]
\centering
\footnotesize
\setlength{\tabcolsep}{6pt}
\setlength{\extrarowheight}{2pt}
\begin{tabular}{lcclcccc}
\toprule
\textbf{Model} & \textbf{w/f} &
\textbf{COMET $\uparrow$} & \textbf{BLEU $\uparrow$} & \textbf{StreamLAAL $\downarrow$} & \textbf{StreamLAAL\textsubscript{CA} $\downarrow$} &
\textbf{NE $\downarrow$} & \textbf{RTF $\downarrow$} \\
\midrule 
\multirow{4}{*}{SeamlessM4T v2 large}
 & 8 & 0.7016 & 21.72 & 3.29 & 4.14 & 0.7062 & 0.4417 \\
& 10 & 0.7089 & 23.31 & 3.57 & 4.55 & 0.8455 & 0.4505 \\
& 12 & 0.7074 & 22.53 & 4.44 & 5.52 & 0.9516 & 0.4903 \\
& 14 & 0.7005 & 20.02 & 6.35 & 7.58 & 1.0316 & 0.6101 \\
\midrule 
\multirow{4}{*}{SeamlessM4T v1 medium}
 & 8  & 0.7174 & 24.51 & 2.42 & 2.80 & 0.8586 & 0.1770 \\
& 10 & 0.7359 & 25.39 & 2.92 & 3.38 & 1.0215 & 0.2101 \\
& 12 & 0.7439 & 25.70 & 3.53 & 4.07 & 1.1778 & 0.2432 \\
& 14 & 0.7469 & 25.65 & 4.26 & 4.90 & 1.3401 & 0.2878 \\
\bottomrule
\end{tabular}
\caption{Quality (COMET, BLUE), latency (StreamLAAL, StreamLAAL\_CA), flickering (NE), and computational cost (RTF) of the sliding window processor using SeamlessM4T v2 large and SeamlessM4T v1 medium, averaged across the 8 language pairs of MuST-C.}
\label{tab:seamless_comp}
\end{table*}

\twocolumn

\begin{table*}[htbp]
\centering
\resizebox{\textwidth}{!}{%
\begin{tabular}{@{}clccccccc@{}}
\toprule
\textbf{Speech P.}  & \textbf{Model} & \textbf{w/f} &
\textbf{COMET $\uparrow$} & \textbf{BLEU $\uparrow$} & \textbf{StreamLAAL $\downarrow$} & \textbf{StreamLAAL\textsubscript{CA} $\downarrow$} &
\textbf{NE $\downarrow$} & \textbf{RTF $\downarrow$} \\
\midrule
\multicolumn{9}{c}{\cellcolor{gray!15}\textbf{en-de}} \\
\midrule
\multirow{8}{*}{\rotatebox{90}{\textcolor{magenta}{\texttt{incremental}}}} & \multirow{4}{*}{SeamlessM4T} & 2 & 0.6535 & 19.48 & 1.49 & 1.67 & 0.0000 & 0.1625 \\
 & & 4 & 0.6723 & 22.17 & 1.78 & 1.98 & 0.0000 & 0.1780 \\
 & & 6 & 0.6799 & 22.86 & 1.96 & 2.16 & 0.0000 & 0.1906 \\
 & & 8 & 0.6859 & 23.82 & 2.31 & 2.51 & 0.0000 & 0.2009 \\
\cmidrule(l){2-9}
 & \multirow{4}{*}{Canary} & 2 & 0.6746 & 17.43 & 2.97 & 3.52 & 0.0000 & 0.4689 \\
 & & 4 & 0.6814 & 17.40 & 3.07 & 3.72 & 0.0000 & 0.5221 \\
 & & 6 & 0.6898 & 17.99 & 3.53 & 4.12 & 0.0000 & 0.5097 \\
 & & 8 & 0.6906 & 18.01 & 3.84 & 4.54 & 0.0000 & 0.5825 \\
\midrule
\multirow{8}{*}{\rotatebox{90}{\textcolor{blue}{\texttt{retranslation}}}} & \multirow{4}{*}{SeamlessM4T} & 8  & 0.6014 & 18.79 & 3.07 & 3.43 & 0.6830 & 0.1757 \\
 & & 10 & 0.6146 & 18.07 & 3.80 & 4.25 & 0.8009 & 0.2133 \\
 & & 12 & 0.6183 & 17.76 & 4.47 & 5.01 & 0.9323 & 0.2474 \\
 & & 14 & 0.6161 & 16.16 & 5.58 & 6.23 & 0.9987 & 0.2858 \\
\cmidrule(l){2-9}
 & \multirow{4}{*}{Canary} & 8  & 0.6855 & 26.01 & 2.68 & 2.99 & 0.7096 & 0.1571 \\
 & & 10 & 0.7071 & 25.67 & 3.21 & 3.66 & 0.8658 & 0.2135 \\
 & & 12 & 0.7490 & 30.19 & 3.52 & 4.08 & 1.0960 & 0.2595 \\
 & & 14 & 0.7593 & 30.02 & 4.06 & 4.70 & 1.2239 & 0.2918 \\
\midrule
\multicolumn{9}{c}{\cellcolor{gray!15}\textbf{en-it}} \\
\midrule
\multirow{8}{*}{\rotatebox{90}{\textcolor{magenta}{\texttt{incremental}}}} & \multirow{4}{*}{SeamlessM4T} & 2 & 0.7198 & 27.59 & 1.47 & 1.64 & 0.0000 & 0.1474 \\
 & & 4 & 0.7420 & 31.84 & 1.68 & 1.86 & 0.0000 & 0.1595 \\
 & & 6 & 0.7496 & 33.56 & 1.90 & 2.09 & 0.0000 & 0.1689 \\
 & & 8 & 0.7585 & 34.79 & 2.22 & 2.41 & 0.0000 & 0.1784 \\
\cmidrule(l){2-9}
 & \multirow{4}{*}{Canary} & 2 & 0.7389 & 23.55 & 2.88 & 3.54 & 0.0000 & 0.5525 \\
 & & 4 & 0.7448 & 25.71 & 2.84 & 3.47 & 0.0000 & 0.5083 \\
 & & 6 & 0.7402 & 24.70 & 2.10 & 2.86 & 0.0000 & 0.5554 \\
 & & 8 & 0.7614 & 26.60 & 3.11 & 3.64 & 0.0000 & 0.4893 \\
\midrule
\multirow{8}{*}{\rotatebox{90}{\textcolor{blue}{\texttt{retranslation}}}} & \multirow{4}{*}{SeamlessM4T} & 8  & 0.7046 & 31.38 & 3.07 & 3.40 & 0.7313 & 0.1614 \\
 & & 10 & 0.7137 & 31.06 & 3.65 & 4.07 & 0.9122 & 0.1901 \\
 & & 12 & 0.7195 & 30.61 & 4.55 & 5.04 & 1.0497 & 0.2218 \\
 & & 14 & 0.7140 & 29.06 & 5.49 & 6.05 & 1.1904 & 0.2561 \\
\cmidrule(l){2-9}
 & \multirow{4}{*}{Canary} & 8  & 0.7811 & 37.96 & 2.56 & 2.90 & 0.6294 & 0.1673 \\
 & & 10 & 0.7888 & 37.59 & 3.03 & 3.40 & 0.7740 & 0.1824 \\
 & & 12 & 0.8145 & 43.18 & 3.43 & 3.94 & 1.0217 & 0.2376 \\
 & & 14 & 0.8186 & 43.81 & 3.80 & 4.38 & 1.1862 & 0.2701 \\
\midrule
\multicolumn{9}{c}{\cellcolor{gray!15}\textbf{en-zh}} \\
\midrule
\multirow{4}{*}{\rotatebox{90}{\textcolor{magenta}{\texttt{\small incremental}}}} & \multirow{4}{*}{SeamlessM4T} & 2 & 0.6845 & 27.35 & 3.18 & 3.50 & 0.0000 & 0.2121 \\
 & & 4 & 0.6957 & 28.78 & 3.41 & 3.68 & 0.0000 & 0.1884 \\
 & & 6 & 0.6992 & 28.78 & 3.64 & 3.91 & 0.0000 & 0.1942 \\
 & & 8 & 0.6978 & 28.85 & 3.39 & 3.64 & 0.0000 & 0.1899 \\
\midrule
\addlinespace[4pt]
\multirow{4}{*}{\rotatebox{90}{\textcolor{blue}{\texttt{\small retranslation}}}} & \multirow{4}{*}{SeamlessM4T} & 8  & 0.6295 & 25.02 & 3.29 & 3.59 & 0.7956 & 0.1423 \\
 & & 10 & 0.6234 & 24.62 & 4.14 & 4.51 & 1.0344 & 0.1690 \\
 & & 12 & 0.6335 & 24.87 & 5.17 & 5.59 & 1.1488 & 0.1942 \\
 & & 14 & 0.6307 & 23.13 & 6.33 & 6.85 & 1.2302 & 0.2215 \\
\addlinespace[4pt]
\bottomrule
\end{tabular}%
}
\caption{MCIF numerical results across speech processors (StreamAtt for \textcolor{magenta}{incremental}, and Sliding Window for \textcolor{blue}{retranslation}), language pairs, and models.}
\label{tab:mcif-results}
\end{table*}
\end{document}